\documentclass[11pt]{article}

\usepackage[preprint]{acl}
\usepackage{stfloats}
\usepackage{placeins}
\usepackage{booktabs}
\usepackage{times}
\usepackage{latexsym}

\usepackage[T1]{fontenc}

\usepackage[utf8]{inputenc}

\usepackage{microtype}

\usepackage{inconsolata}

\usepackage{graphicx}
\graphicspath{{./}{../}{latex/}}

\title{The Visual Bottleneck: Sparse-Frame Adaptation of MLLMs for Joint Spatial-Temporal Video Grounding}

\author{Jiameng Zhang \\
  University of Zurich \\
  \texttt{jiameng.zhang@uzh.ch} \\\And
  Srikanth Madikeri \\
  University of Zurich \\
  \texttt{srikanth.madikeriraghunathan@uzh.ch} \\}

\begin{document}
\maketitle

\begin{abstract}
Large-scale video platforms process millions of uploads hourly, requiring
moderation systems that can localize when and where policy violations occur within
each video. Processing every frame is infeasible at scale, so systems are
constrained to sparse inputs of 8 to 16 frames per video. \citep{miech2019howto100m, tang2024videollmsurvey}
Yet state-of-the-art multimodal large language models (MLLMs) are pretrained on
dense sequences of hundreds of frames, creating a fundamental mismatch between
training and deployment conditions. This mismatch causes severe performance
collapse: the Qwen3-VL 8B model drops from 56.0\% to 22.3\% temporal mIoU when
frames are reduced to 16, a 60.2\% relative degradation.
 
We present a systematic empirical study of training strategies to close this gap
for spatial-temporal video grounding. Our results suggest that visual feature
extraction is the dominant bottleneck under sparse-frame inputs. Adapting only the
final three ViT layers, 4\% of total parameters, achieves 68.8\% temporal mIoU
and surpasses a zero-shot 8B model using dense inputs by 12.8 points. Language
model fine-tuning, by contrast, offers negligible or negative returns. A
boundary-aware sampling strategy, \textsc{Hybrid16}, further improves temporal
mIoU by 26 points over uniform sampling when temporal boundaries are available. We
conclude that for sparse-frame video grounding, training strategy dominates model
scale: a fine-tuned 2B model consistently outperforms a zero-shot 8B model, with
or without dense frame access.
\end{abstract}
 
\section{Introduction}
 
Spatial-temporal video grounding requires models to localize \textit{when} and
\textit{where} actions occur in videos given natural language queries
\citep{gao2017tall, hendricks2017localizing}. This capability is central to
industrial applications such as content moderation, video retrieval, and automated
surveillance, where systems must identify both the temporal extent and spatial
location of query-relevant events at scale.

\begin{figure*}[!t]
    \centering
    \includegraphics[width=\textwidth]{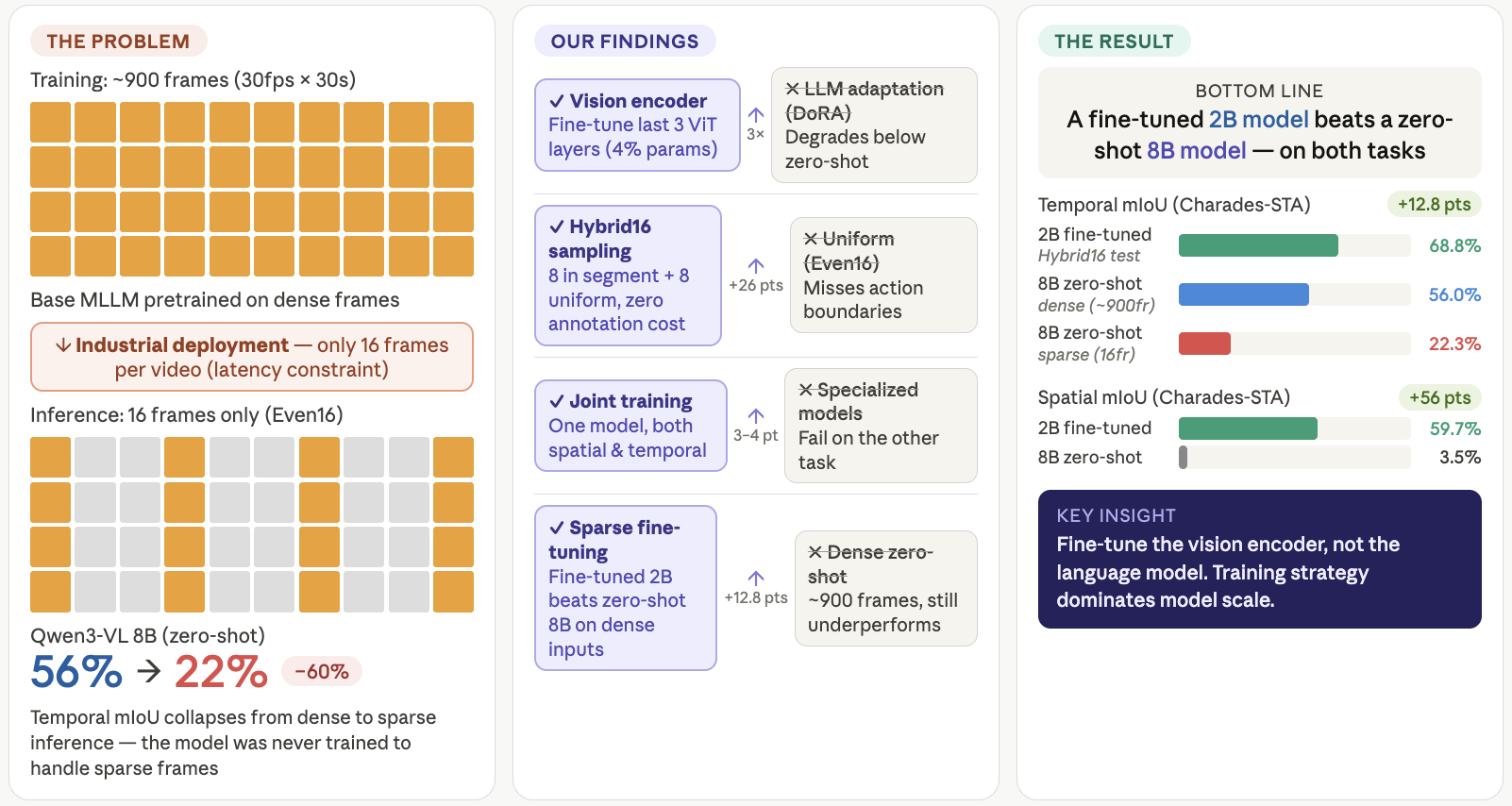}
    \caption{Overview of the train-dense/infer-sparse mismatch 
    and our proposed solutions. Left: the Qwen3-VL 8B model 
    drops from 56.0\% to 22.3\% temporal mIoU when frames are 
    reduced from hundreds to 16. Center: our four key findings. 
    Right: a fine-tuned 2B model outperforms a zero-shot 8B model 
    on both temporal and spatial grounding.}
    \label{fig:overview}
\end{figure*}
 
A critical deployment reality has received insufficient attention in prior work.
Industrial video systems are constrained to sparse inputs of 8 to 16 frames per
video to satisfy latency and throughput requirements, yet state-of-the-art
multimodal large language models (MLLMs) are pretrained on dense sequences of
hundreds of frames. The resulting mismatch between training and deployment
conditions causes severe performance degradation that makes existing models
impractical for real-world use without targeted adaptation.
 
Prior work on video temporal grounding has largely focused on designing new model
architectures and visual aggregation modules \citep{wasim2024videogroundingdino,
liu2024r2tuning, xiong2024structured, gu2024context}, answering the question of
how to build a better grounding model from scratch. We instead address a
complementary and more practical question: given a strong pretrained MLLM, what
training strategy is most effective for sparse-frame spatial-temporal grounding?
This question directly reflects the situation faced by practitioners, who must
adapt existing models to deployment constraints rather than design new architectures.

We investigate this question through systematic experiments on Qwen3-VL models and
Charades-STA \citep{gao2017tall, sigurdsson2016charades}. Since Charades-STA provides only temporal
annotations, we construct a spatial annotation extension using GroundingDINO
\citep{liu2023groundingdino}, yielding 3,951 bounding box annotations with 85\%
acceptance rate under manual inspection. These annotations provide pseudo-labeled
spatial supervision rather than exhaustively verified gold labels. Our experiments compare vision encoder
versus language model adaptation, joint versus single-task training, and multiple
frame sampling strategies under realistic sparse-input constraints.
 
Our findings challenge common assumptions about MLLM adaptation. Fine-tuning
language model attention layers via DoRA \citep{liu2024dora} degrades performance
below zero-shot baselines, while adapting only the final three ViT layers achieves
68.8\% temporal mIoU and surpasses a zero-shot 8B model using dense inputs by
12.8 points. This suggests that the pretrained language model already provides
substantial temporal-query understanding, while sparse visual feature extraction is the main bottleneck for
sparse-frame grounding. A boundary-aware sampling strategy that reallocates frames
from existing temporal annotations to action segments further improves temporal
mIoU by 26 points over uniform sampling in annotation-aware settings.
 
These findings yield four contributions. We identify the training and deployment
mismatch as a critical but underaddressed challenge for industrial video grounding
and provide the first systematic analysis of training strategies to close this gap.
We show that adapting only 4\% of model parameters on a single GPU suffices to
outperform models three times larger under zero-shot dense inference. We propose
\textsc{Hybrid16}, a boundary-aware sampling strategy for pre-annotated archives
and re-inspection pipelines. We contribute a spatial annotation extension for Charades-STA
and a paired evaluation framework using Peak and Robust Performance Scores for
comprehensive spatial-temporal assessment.

\section{Related Work}

\paragraph{Video Temporal Grounding.}
Early approaches to temporal grounding relied on cross-modal attention mechanisms to align natural language queries with video features \citep{gao2017tall, hendricks2017localizing}. Subsequent work focused on designing specialized architectures, including proposal-based methods that generate candidate temporal segments and rank them by query relevance, and proposal-free methods that directly regress temporal boundaries. More recent work has explored integrating temporal grounding with spatial localization. \citet{wasim2024videogroundingdino} extend GroundingDINO to spatio-temporal video grounding in an open-vocabulary setting, while \citet{xiong2024structured} propose structured video-language modeling with joint temporal grouping and spatial grounding. \citet{liu2024r2tuning} address efficient image-to-video transfer learning for temporal grounding. In contrast to these architecture-focused approaches, we investigate training strategy choices for adapting a pretrained MLLM to sparse-frame grounding, a deployment constraint largely overlooked in prior work.

\paragraph{Sparse Frame Processing for Video Grounding.}
Industrial-scale video platforms process millions of uploads
requiring real-time analysis, making dense frame processing
computationally prohibitive \citep{miech2019howto100m}.
Prior work has examined how frame sampling strategies influence
temporal grounding performance \citep{zhu2023rethinking}, and
recent training-free approaches demonstrate that large pretrained
models can localize moments without task-specific supervision
\citep{zheng2024trainingfree}. However, both lines of work
assume access to dense frame sequences at inference time.
The performance collapse that arises when models pretrained
on dense inputs are deployed with sparse frames remains
unaddressed. We provide the first systematic study of training
strategies specifically designed to close this train-deploy gap.

\paragraph{Multimodal Large Language Models for Video.}
MLLMs have emerged as a unified framework for vision-language tasks, combining pretrained vision encoders with large language models through lightweight connectors \citep{alayrac2022flamingo, li2023blip2}. Video-LLaMA \citep{zhang2023videollama, maaz2024videochatgpt} and Video-ChatGPT 
\citep{maaz2024videochatgpt} extend this paradigm to video
inputs through instruction tuning on visual and auditory content. Recent models such as Qwen3-VL \citep{qwen3vl2025} demonstrate strong performance on video understanding benchmarks through multi-level visual feature fusion and extended context windows. Recent work specifically
targeting temporal grounding with video LLMs, such as VTimeLLM
\citep{huang2024vtimellm}, further demonstrates the importance of
boundary-aware training for precise temporal localization. However, these models are pretrained on dense frame sequences, creating a fundamental mismatch with industrial deployment constraints that limit inputs to 8--16 frames per video. We directly address this train-deploy mismatch through targeted adaptation strategies. 

\paragraph{Parameter-Efficient Fine-Tuning.}
PEFT methods reduce the cost of adapting large pretrained models by updating only a small subset of parameters. LoRA \citep{hu2021lora} decomposes weight updates into low-rank matrices, while DoRA \citep{liu2024dora} further separates magnitude and direction components for more expressive adaptation. These methods have been widely applied to language model adaptation, but their effectiveness for video grounding remains underexplored. Our experiments demonstrate that applying DoRA to language model attention layers degrades performance below zero-shot baselines, while selective vision encoder fine-tuning achieves substantially stronger results, pointing to visual feature extraction rather than language reasoning as the primary bottleneck.

\section{Dataset and Annotation}

\subsection{Charades-STA}

Charades-STA \citep{gao2017tall} is a standard benchmark for video temporal
grounding, containing 12,408 training and 3,720 test query-segment pairs drawn
from 9,849 indoor activity videos. Each query consists of a natural language
description paired with temporal boundary annotations marking the relevant action
segment. Video durations average approximately 30 seconds, providing sufficient
temporal context while remaining computationally tractable for sparse-frame
processing. Charades-STA is particularly suitable for this study because its
controlled indoor environment and dense temporal annotations allow focused
evaluation of spatial-temporal grounding under realistic computational constraints.

\subsection{Spatial Annotation Pipeline}

Since Charades-STA provides only temporal annotations, we extend the dataset with
spatial bounding box annotations to enable joint spatial-temporal training and
evaluation. We use GroundingDINO \citep{liu2023groundingdino}, an open-vocabulary
object detection model, to automatically generate spatial annotations from the
existing natural language queries.

The pipeline processes 131,000 frames extracted at 1 fps from Charades-STA videos,
using the natural language query associated with each video as the detection prompt.
For example, given the query ``person opens refrigerator,'' GroundingDINO detects
``person'' and ``refrigerator'' in relevant frames and outputs bounding box
coordinates with confidence scores. Detections with confidence below 0.3 are
filtered out. To prevent redundant training examples, we retain one bounding box
per query-video pair, selected by detection confidence and temporal proximity to
the annotated action segment. Overrepresented object categories are subsampled to
prevent class imbalance. This process yields 3,951 spatial annotations covering
diverse indoor object categories.

To assess annotation quality, we manually inspect a random sample of 100
annotations. An annotation is accepted if the predicted bounding box reasonably
localizes the object referred to in the query. This inspection yields an 85\%
acceptance rate, indicating that the pipeline produces useful spatial supervision
for training and diagnostic evaluation. Because the full spatial split is not
exhaustively human verified, we interpret spatial metrics as evaluation against
high-confidence pseudo labels rather than fully gold spatial annotations.

\subsection{Training Data}

The 3,951 spatial annotations are combined with temporal samples to form five
training configurations, as detailed in Section~\ref{sec:setup}. Spatial instances
use single frames at 480$\times$360 resolution with bounding box annotations in
$[x_{\min}, y_{\min}, x_{\max}, y_{\max}]$ format. Temporal instances use 16
frames with ground truth start and end timestamps. Two non-overlapping test sets
of 1,000 instances each are constructed from the Charades-STA test split using
Even16 and Hybrid16 sampling respectively, ensuring no overlap with training data.

\section{Experimental Setup}
\label{sec:setup}

\subsection{Model and Training Strategy}

We use Qwen3-VL 2B \citep{qwen3vl2025} as our foundation model. The architecture
consists of a Vision Transformer encoder \citep{dosovitskiy2021vit} (24 layers, approximately 300M parameters),
a text Transformer decoder (28 layers, approximately 1.5B parameters), and a
DeepStack multimodal connector that fuses features from multiple encoder depths.
The model has been pretrained on large-scale vision-language data including
video understanding tasks.

Our training strategy freezes the text Transformer entirely to preserve pretrained
language understanding, and also freezes early vision encoder layers (blocks 0--20)
to retain general visual features such as edges, textures, and local patterns.
We make trainable only the final three ViT blocks (21--23), the Merger and
DeepStack Merger projection modules, and the patch embedding and positional
encoding layers. This yields approximately 70M trainable parameters, or 4\% of
the total 1.75B parameters, enabling training on a single NVIDIA H100 (80GB GPU).
Table~\ref{tab:params} summarizes the parameter distribution.

\begin{table}[h]
\centering
\small
\renewcommand{\arraystretch}{1.15}
\begin{tabular}{lcc}
\toprule
\textbf{Component} & \textbf{Total} & \textbf{Trainable} \\
\midrule
Text Transformer & 1.5B & 0 \\
Vision Encoder (Blocks 0--20) & 180M & 0 \\
Vision Encoder (Blocks 21--23) & 45M & 45M \\
Merger Modules & 20M & 20M \\
Embedding Layers & 5M & 5M \\
\midrule
Total & 1.75B & 70M (4\%) \\
\bottomrule
\end{tabular}
\caption{Parameter distribution across model components.}
\label{tab:params}
\end{table}

\subsection{Frame Sampling Strategies}

We compare three frame sampling strategies that balance temporal coverage and
computational efficiency.

\textbf{FPS1} extracts one frame per second, yielding approximately 30 frames per
video. This provides dense temporal coverage but results in variable sequence
lengths, and is used only for spatial annotation instances.

\textbf{Even16} selects 16 frames uniformly across the entire video duration,
ensuring fixed input size and consistent memory usage. This strategy reflects
common practice in video understanding systems but may undersample action regions
when the target event occupies a small portion of the video timeline.

\textbf{Hybrid16} addresses this limitation by allocating 8 frames to the
annotated action segment and 8 frames uniformly across the full video. This
design concentrates coverage at temporal boundaries where state transitions are
most informative, while retaining global context, at zero additional annotation
cost beyond the temporal labels already present in Charades-STA. Hybrid16 sampling requires temporal annotations at inference time and is 
therefore suited to pre-annotated video archives and re-inspection pipelines; 
for deployment scenarios where no temporal annotations are available, Even16 
provides a fully annotation-free alternative with stable performance.

Table~\ref{tab:sampling} illustrates the difference between strategies for a
representative video.

\begin{table}[h]
\centering
\small
\renewcommand{\arraystretch}{1.15}
\begin{tabular}{lccc}
\toprule
\textbf{Strategy} & \textbf{In Segment} & \textbf{Outside} & \textbf{Total} \\
\midrule
FPS1  & 8  & 23 & 31 \\
Even16   & 3  & 13 & 16 \\
Hybrid16 & 10 & 6  & 16 \\
\bottomrule
\end{tabular}
\caption{Frame distribution for a 30-second video with action segment at
11.2--16.7s.}
\label{tab:sampling}
\end{table}

\begin{figure*}[h]
    \centering
    \includegraphics[width=1.8\columnwidth]{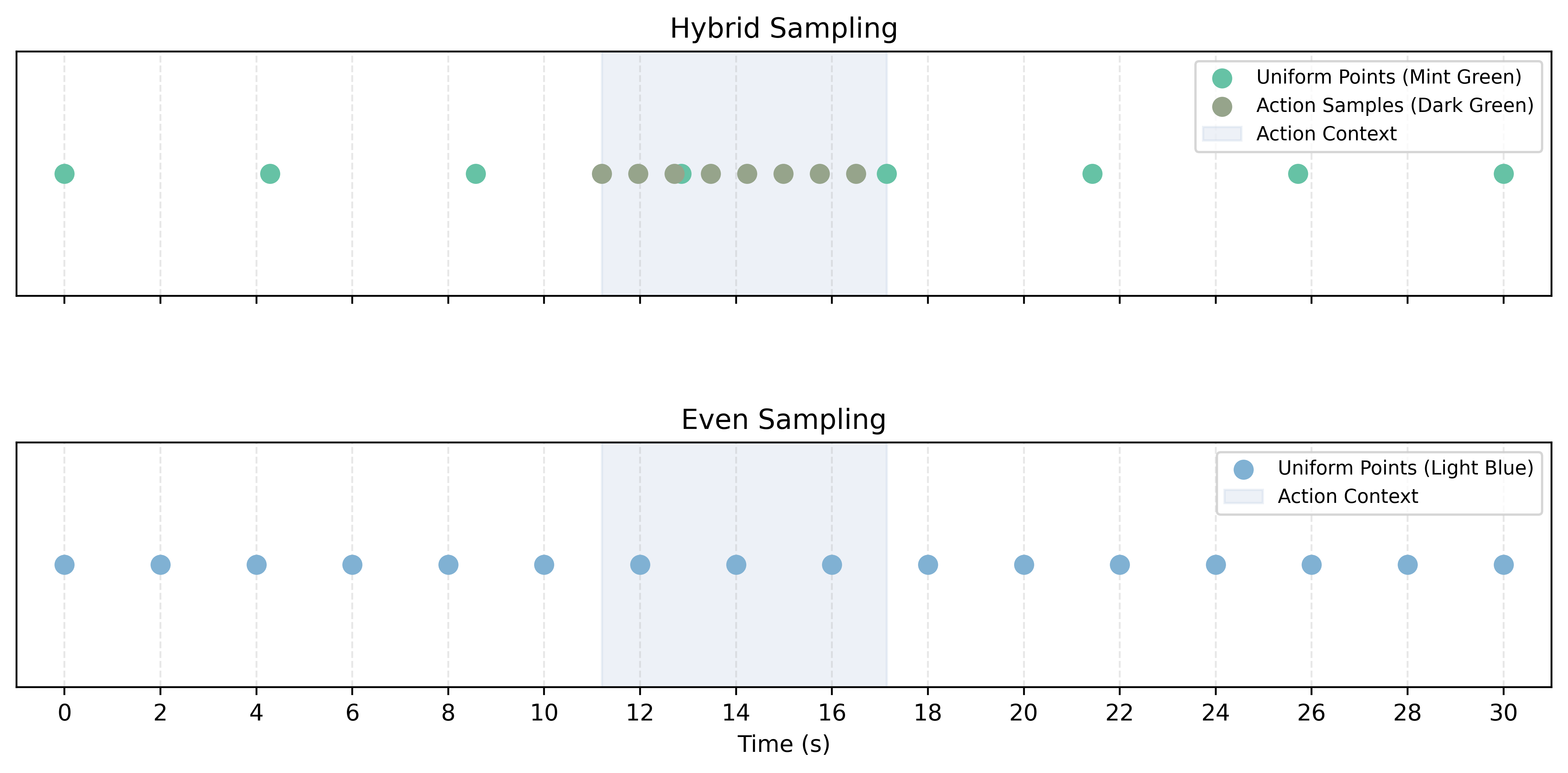}
    \caption{Comparison of \textsc{Hybrid16} and Even16 sampling 
    strategies for a 30-second video with action segment at 
    11.2--16.7s. \textsc{Hybrid16} concentrates 8 frames within 
    the action segment (dark green) while distributing 8 frames 
    uniformly (mint green). Even16 distributes all 16 frames 
    uniformly across the video.}
    \label{fig:sampling}
\end{figure*}

\subsection{Training Configurations}

We design five training configurations to evaluate the impact of data composition
and sampling strategy. Each configuration is trained independently on Qwen3-VL 2B
and evaluated on both Even16 and Hybrid16 test sets to assess generalization across
frame distributions. Table~\ref{tab:configs} summarizes the configurations.

\begin{table}[h]
\centering
\small
\renewcommand{\arraystretch}{1.15}
\begin{tabular}{lccc}
\toprule
\textbf{Config} & \textbf{Size} & \textbf{Sampling} & \textbf{Ann.} \\
\midrule
BBx-only    & 3,951 & FPS1            & Spatial \\
Even-only   & 7,000 & Even16          & Temporal \\
Hybrid-only & 7,000 & Hybrid16        & Temporal \\
Evenbbx     & 6,500 & FPS1 + Even16   & Both \\
Hybridbbx   & 6,500 & FPS1 + Hybrid16 & Both \\
\bottomrule
\end{tabular}
\caption{Training configurations. Joint configurations (Evenbbx, Hybridbbx)
combine 2,500 spatial and 4,000 temporal instances.}
\label{tab:configs}
\end{table}

All models are trained for 3 epochs using AdamW with learning rate $5 \times
10^{-6}$, BF16 mixed precision, gradient checkpointing, and cosine annealing
with 10\% linear warmup. Images are resized to fit within 256$\times$256 to
512$\times$512 pixels while preserving aspect ratio. Training was conducted on
a single NVIDIA H100 GPU using the LLaMA-Factory framework \citep{zheng2024llamafactory}.

\subsection{Evaluation Metrics}

Temporal grounding performance is measured using mean Intersection over Union
(mIoU) and Recall at IoU thresholds $R@\{0.3, 0.5, 0.7\}$. Spatial grounding
is evaluated using spatial mIoU and Precision@0.5. To assess joint
spatial-temporal capability, we propose two complementary metrics. The Peak
Performance Score averages the best temporal mIoU across both test sets with
spatial mIoU, measuring achievable performance under favorable conditions. The
Robust Performance Score averages temporal mIoU across both test distributions
with spatial mIoU, quantifying stability under distribution shift. These paired
metrics address the limitation of evaluating spatial and temporal grounding in
isolation, which fails to capture trade-offs between the two dimensions.

\begin{table}[h]
\centering
\small
\renewcommand{\arraystretch}{1.15}
\resizebox{\columnwidth}{!}{%
\begin{tabular}{lcccc}
\toprule
\textbf{Config} & \textbf{Spatial} & \textbf{Peak} & \textbf{Robust} \\
\midrule
Hybridbbx   & 0.60 & \textbf{0.64} & \textbf{0.55} \\
Evenbbx     & 0.55 & 0.51 & 0.50 \\
Hybrid-only & 0.00 & 0.37 & 0.24 \\
Even-only   & 0.01 & 0.25 & 0.34 \\
BBx-only    & 0.60 & 0.33 & 0.33 \\
\bottomrule
\end{tabular}}
\caption{Joint spatial-temporal evaluation using Peak and Robust 
Performance Scores. Peak = (best temporal + spatial mIoU) / 2. 
Robust = (average temporal + spatial mIoU) / 2.}
\label{tab:joint}
\end{table}

\section{Results}
\label{sec:results}

\subsection{Vision Encoder vs. Language Model Adaptation}

We first investigate which model component is most critical to adapt for
sparse-frame grounding. As a baseline, we apply DoRA \citep{liu2024dora} with
rank 64 to all language model attention layers while keeping the vision encoder
frozen. This language-only adaptation degrades temporal mIoU to 15--16\%, below
the zero-shot baseline of 21--22\%, demonstrating that modifying language
components disrupts pretrained temporal query understanding without compensating
benefits. The pretrained language model already handles temporal queries
adequately; further adaptation introduces noise rather than improvement.

Adapting the vision encoder produces the opposite effect. Fine-tuning only the
final three ViT layers achieves 68.8\% temporal mIoU on matched test
distributions and over 55\% spatial mIoU, more than three times higher than
language-only adaptation. Table~\ref{tab:vision_vs_lm} summarizes the comparison.

\begin{table}[h]
\centering
\small
\renewcommand{\arraystretch}{1.15}
\begin{tabular}{llcc}
\toprule
\textbf{Strategy} & \textbf{Params} & \textbf{Even16} & \textbf{Hybrid16} \\
\midrule
Zero-shot 8B   & None         & 0.21 & 0.22 \\
DoRA (LM only) & LM attention & 0.15 & 0.16 \\
Vision-focused & ViT 21--23   & 0.47 & \textbf{0.69} \\
\bottomrule
\end{tabular}
\caption{Temporal mIoU for language model adaptation vs.\ vision encoder
adaptation. Vision-focused training uses the Evenbbx and Hybridbbx
configurations on their respective matched test sets.}
\label{tab:vision_vs_lm}
\end{table}

This asymmetry is further supported by zero-shot scaling results. Expanding the
language model from 28 to 36 layers (2B to 4B, same vision encoder) yields only
5.1 points improvement on Hybrid16. Expanding the vision encoder from 24 to 27
layers (4B to 8B, same LM depth) yields 8.0 points despite adding fewer layers.
Vision encoder capacity contributes more to grounding performance than language
model capacity, identifying visual feature extraction as the primary bottleneck.

\subsection{Sparse Fine-tuning vs. Dense Zero-shot}

Zero-shot models rely on dense frame processing to achieve competitive performance.
The Qwen3-VL 8B model achieves 56.0\% temporal mIoU using all available frames
but drops to 22.3\% when limited to 16 frames, a 60\% relative collapse. Without
task-specific training, models cannot identify which sparse frames contain
action-relevant information.

Fine-tuning reverses this relationship entirely. As shown in Table~\ref{tab:dense_vs_sparse},
the fine-tuned 2B model achieves 68.8\% temporal mIoU using 16 frames, surpassing
the zero-shot 8B model under dense processing by 12.8 points and under sparse
processing by 46.5 points. At stricter evaluation thresholds, the advantage
becomes more pronounced: the fine-tuned model achieves 64.0\% Recall@0.7 on
Hybrid16 compared to 4.0\% for zero-shot 8B. These results demonstrate that
targeted adaptation can outweigh model scale under sparse-frame constraints.

\begin{table}[h]
\centering
\small
\renewcommand{\arraystretch}{1.15}
\resizebox{\columnwidth}{!}{%
\begin{tabular}{llccc}
\toprule
\textbf{Model} & \textbf{Frames} & \textbf{mIoU} & \textbf{R@0.5} & \textbf{R@0.7} \\
\midrule
Zero-shot 8B      & All frames    & 0.56 & --   & --   \\
Zero-shot 8B      & 16 (Hyb16)    & 0.22 & 0.13 & 0.04 \\
Zero-shot 2B      & 16 (Hyb16)    & 0.09 & 0.03 & 0.01 \\
Fine-tuned 2B$^\dagger$ & 16 (Hyb16) & \textbf{0.69} & \textbf{0.74} & \textbf{0.64} \\
\bottomrule
\end{tabular}}
\caption{Temporal grounding performance comparing dense zero-shot inference
against sparse fine-tuned models. $^\dagger$Hybridbbx configuration.}
\label{tab:dense_vs_sparse}
\end{table}

\subsection{Frame Sampling Strategy}

Sampling strategy creates strong dependencies between training and test
distributions. Models perform best when test conditions match training conditions.
Hybridbbx achieves 68.8\% mIoU on Hybrid16 test but drops to 33.6\% on Even16
test, a 51\% relative decline. Evenbbx shows the opposite pattern, achieving
47.4\% on Even16 and 42.8\% on Hybrid16, a more stable 10\% decline under
distribution shift.

This asymmetry reflects different learned representations. Hybrid16 training
concentrates frames near action boundaries, teaching the model to recognize
fine-grained temporal transitions. These specialized features prove highly
effective under aligned test conditions but transfer poorly when frame
distributions change. Even16 training exposes the model to frames from all
temporal positions, developing more general representations that maintain
consistent performance across distributions at the cost of peak capability.

For spatial grounding, sampling strategy has a much smaller effect. Hybridbbx
achieves 59.7\% spatial mIoU compared to 55.3\% for Evenbbx, a 4.4 point
difference, versus the 26 point gap observed for temporal grounding on matched
distributions. Spatial grounding depends on object localization within individual
frames rather than temporal frame ordering, explaining its relative insensitivity
to sampling strategy.

\subsection{Joint vs. Specialized Training}

Specialized models achieve the highest scores on their target tasks. Hybrid-only
reaches 72.7\% temporal mIoU on Hybrid16, outperforming Hybridbbx by 3.9 points.
BBx-only achieves 60.0\% spatial mIoU, exceeding Hybridbbx by 0.3 points.
However, specialized models fail completely on the complementary task: Hybrid-only
achieves 0.34\% spatial mIoU and BBx-only achieves 6.0\% temporal mIoU.

To assess overall joint capability, we apply the Peak and Robust Performance
Scores introduced in Section~\ref{sec:setup}. As shown in Table~\ref{tab:joint},
Hybridbbx achieves the highest scores on both metrics (0.64 peak, 0.55
robust), demonstrating that joint training provides balanced capability across
both tasks. The small gap between peak and robust scores (0.09) indicates
consistent performance across test conditions. Specialized configurations score
poorly on both metrics despite strong single-task performance, as near-zero
performance on one dimension drags down the combined score. The performance
trade-off from joint training is modest: 5.4\% relative loss for temporal
grounding and 0.5\% for spatial grounding, a reasonable cost for maintaining
full dual-task capability within a single model.

\section{Discussion}

\paragraph{Why vision encoder adaptation succeeds.}
Temporal grounding requires recognizing subtle visual cues that indicate action
boundaries, state changes, and temporal transitions. General-purpose vision
encoders trained on image-text alignment do not capture these patterns well.
The final ViT layers encode high-level semantic features that are most amenable
to task-specific adaptation, while early layers capture low-level visual
primitives that transfer well across tasks and benefit from being frozen.
Training only 4\% of parameters is sufficient because the bottleneck is
concentrated in these upper layers, not distributed across the entire model.
This finding aligns with concurrent observations in image understanding
\citep{bi2024llavasterring, chen2024internvl}, suggesting it reflects a structural property of
MLLM architectures rather than a dataset-specific artifact.

\paragraph{Why language model adaptation fails.}
Modern large language models acquire extensive knowledge of temporal concepts
and action semantics during pretraining. They already understand phrases such
as ``person starts to drink'' or ``someone picks up a phone'' and can reason
about temporal relationships between events. Adapting these representations
with DoRA does not address the actual challenge, which is connecting linguistic
concepts to visual observations in sparse frame sequences. Modifying language
components instead disrupts existing capabilities, explaining the performance
degradation below zero-shot baselines.

\paragraph{Hybrid sampling and action boundaries.}
Human actions have natural boundaries with concentrated visual transitions at
their start and end points. Hybrid16 exploits this structure by allocating half
its frame budget to the annotated action segment, providing dense coverage of
precisely the frames that carry the most discriminative temporal information.
The large performance gap between Hybrid16 and Even16 on matched distributions
(68.8\% vs. 42.8\%) demonstrates the value of boundary-aware sampling, while
the robustness gap (51\% vs. 10\% decline under distribution shift) reveals the
trade-off between specialization and generalization. For deployment scenarios
where temporal boundaries are already available or can be supplied by an upstream
system, Hybrid16 is the stronger annotation-aware option.
For scenarios with unknown frame distributions, Even16 provides more stable
performance.

\paragraph{Practical implications.}
These findings provide concrete guidance for practitioners adapting MLLMs to
video grounding tasks. Vision encoder fine-tuning should be considered before
language model adaptation, as our experiments indicate that it addresses the dominant bottleneck while
preserving pretrained linguistic capabilities. Hybrid frame sampling is
applicable to datasets and workflows with temporal annotations, such as curated
archives or re-inspection pipelines. Joint spatial-temporal training is recommended
over specialized models for most deployment scenarios, as the modest 3--4
point performance trade-off is outweighed by the operational benefit of
maintaining a single model for both tasks.

\paragraph{Implications for the research community.}
Much prior work on video temporal grounding emphasizes
cross-modal interaction mechanisms, attention-based fusion
modules, and temporal reasoning architectures
\citep{wasim2024videogroundingdino, xiong2024structured}.
These approaches address how to better combine visual and
linguistic information, implicitly assuming that the visual
representations themselves are adequate. Our results challenge
this assumption: in our setting, language adaptation provides negligible gains,
while vision encoder adaptation yields over 200\% relative
improvement. This suggests that visual representation learning
for sparse-frame grounding deserves greater attention as a
research direction in its own right, complementing work on
cross-modal fusion. Targeted adaptation of the vision encoder,
rather than scaling model size or refining fusion mechanisms,
may offer a more efficient path to deployment-ready
spatial-temporal grounding.

\section{Conclusion}

We have presented a systematic empirical study of training strategies for
spatial-temporal video grounding under the sparse-frame constraints of
industrial deployment. Our evidence indicates that sparse visual feature extraction,
rather than language-model adaptation, is the dominant bottleneck in this setting. Adapting
only the final three ViT layers surpasses language model fine-tuning by a
factor of three and outperforms a zero-shot model four times larger. A
boundary-aware frame sampling strategy further improves temporal grounding by
26 points under matched annotation-aware test conditions, and joint spatial-temporal
training achieves balanced capability across both tasks with minimal trade-off.
Together, these results show that targeted adaptation can be more effective than
model scaling for sparse-frame video grounding, offering a practical path to
high-performance deployment without requiring large models or dense frame
processing.

\section{Limitations}
Our study has four primary limitations. First, all experiments 
are conducted on Charades-STA, which features short indoor 
activity videos with controlled environments; whether the 
findings generalize to long-form videos, egocentric recordings, 
or outdoor scenes remains an open question. Second, spatial 
annotations are automatically generated by GroundingDINO and 
validated on a sample of 100 annotations (85\% acceptance 
rate). As a result, spatial results should be interpreted as 
evaluation against high-confidence pseudo labels rather than 
fully human-verified gold annotations; larger-scale human validation 
would further strengthen the reliability of spatial grounding 
evaluation. Third, the 
choice to fine-tune the final three ViT layers was based on 
preliminary experiments rather than systematic ablation; other 
layer selection strategies may yield further improvements. 
Fourth, the comparison between joint and specialized training 
is confounded by differences in training set size, as joint 
configurations use 6,500 instances while specialized models 
use 3,951--7,000; future work should control for this variable 
to isolate the effect of task composition from data quantity. 
We leave cross-dataset validation, layer selection ablation, 
and controlled joint training experiments to future work.

\section{Ethical Considerations}
This work studies video grounding methods that may be used in content moderation,
archive search, and monitoring workflows. Such systems can affect privacy and
fairness when applied to people in videos, especially if model predictions are
used for enforcement or surveillance without human review. We therefore view the
proposed methods as decision-support tools rather than standalone systems for
high-stakes decisions. The spatial annotations used in this study are generated
with an automatic detector and are not exhaustively human verified, so downstream
uses should account for annotation noise and potential detector bias.

\bibliography{references}

\clearpage
\twocolumn
\appendix
\newcommand{\caseimage}[1]{%
\includegraphics[width=0.96\textwidth,trim=0 28pt 0 62pt,clip]{#1}}

\section{Prompt Templates}
\label{app:prompts}

All prompts follow the Qwen3-VL conversation format with
interleaved image and text tokens.

\subsection{Temporal Grounding Prompt}
The temporal grounding prompt provides the model with a list of frame 
timestamps and the corresponding 16 video frames, followed by a natural 
language action query. The model is instructed to output only the start 
and end time in seconds. The timestamps reflect the sampling strategy 
used during training or evaluation: Hybrid16 timestamps concentrate 
near action boundaries, while Even16 timestamps are distributed 
uniformly across the video duration (see Section~\ref{sec:setup} 
for details).

{\small
\begin{verbatim}
User:
Timestamps: [t_1, t_2, ..., t_16].
<image> <image> ... <image>
(16 frames)
Locate the following action: {query}.
Only output the start and end time
in seconds.
Assistant:
{start_time}, {end_time}
\end{verbatim}
}

\subsection{Spatial Grounding Prompt}
The spatial grounding prompt provides the model with a single video 
frame and instructs it to output bounding box coordinates for the 
queried objects. Coordinates are expressed in pixel space as 
$[x_{\min}, y_{\min}, x_{\max}, y_{\max}]$.

{\small
\begin{verbatim}
User:
<image>
Provide bounding box coordinates
for the following objects:
{query}.
Only output coordinates as:
[x_min, y_min, x_max, y_max].
Assistant:
{object} <bbox>
x_min y_min x_max y_max
</bbox>
\end{verbatim}
}

\section{Training Loss Curves}
\label{app:training_loss}

\begin{figure}[htbp]
\centering
\includegraphics[width=\columnwidth]{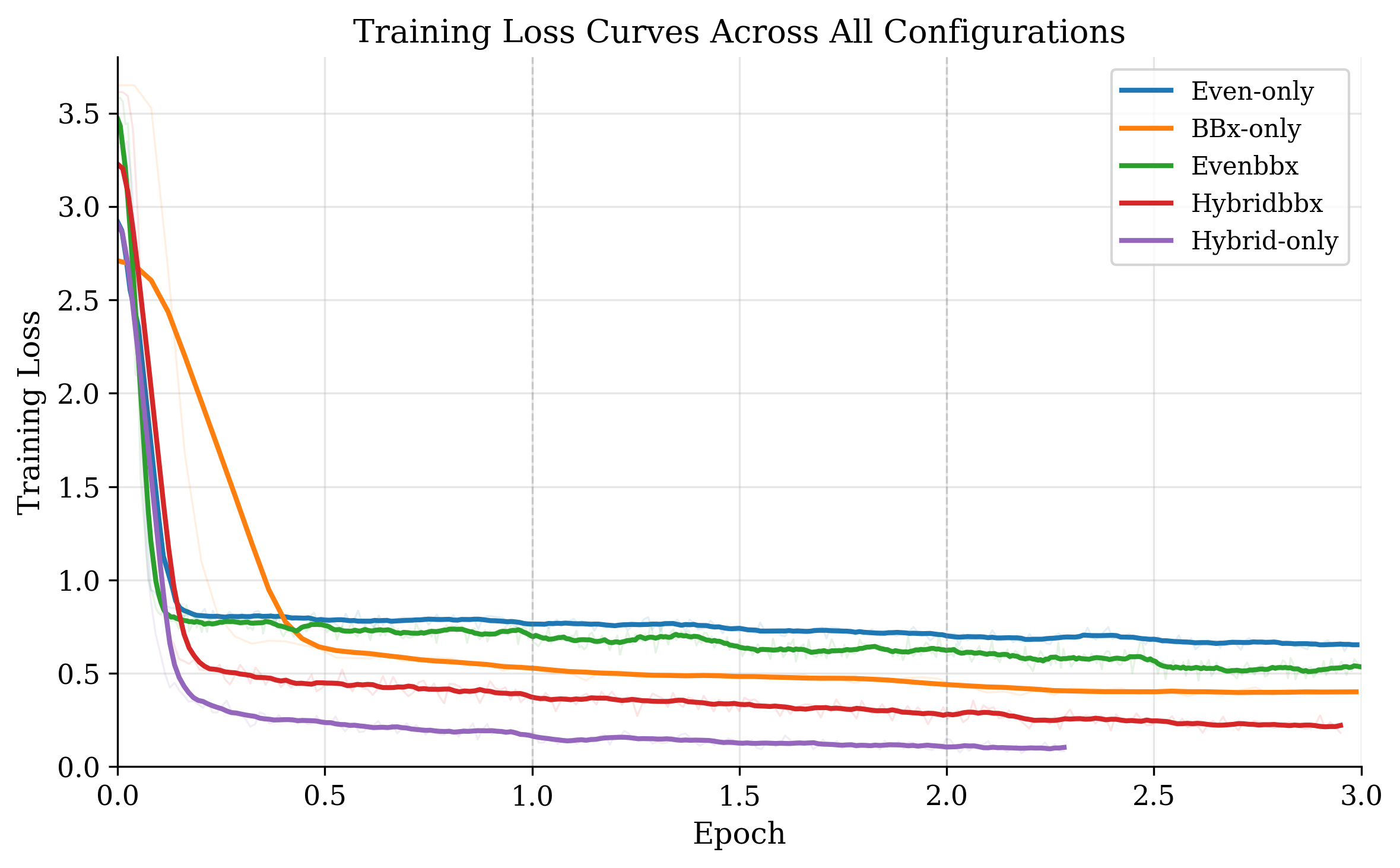}
\caption{Training loss curves for all five configurations. 
Hybrid-only training was stopped early at epoch 2.28 following 
loss plateau.}
\label{fig:training_loss}
\end{figure}

Figure~\ref{fig:training_loss} presents training loss curves for all 
five configurations. All configurations converge smoothly within three 
epochs. BBx-only reaches a lower final loss than temporal 
configurations, reflecting the simpler single-frame spatial grounding 
objective. Hybrid-only was stopped early at epoch 2.28 following loss 
plateau. Dashed vertical lines indicate epoch boundaries.

\section{Complete Experimental Results}

Tables~\ref{tab:full_temporal} and~\ref{tab:full_spatial} present 
complete temporal and spatial grounding results across all training 
configurations and test distributions.

\begin{table*}[htbp]
\centering
\footnotesize
\renewcommand{\arraystretch}{1.0}
\resizebox{0.7\textwidth}{!}{%
\begin{tabular}{llcccc}
\toprule
\textbf{Configuration} & \textbf{Test Set} & \textbf{mIoU} & 
\textbf{R@0.3} & \textbf{R@0.5} & \textbf{R@0.7} \\
\midrule
\multicolumn{6}{l}{\textit{Zero-shot Dense (All Frames)}} \\
Qwen3-VL-2B & All frames & 0.55 & -- & -- & -- \\
Qwen3-VL-4B & All frames & 0.56 & -- & -- & -- \\
Qwen3-VL-8B & All frames & 0.56 & -- & -- & -- \\
\midrule
\multicolumn{6}{l}{\textit{Zero-shot Sparse (16 Frames)}} \\
Qwen3-VL-2B & Even16   & 0.13 & 0.15 & 0.04 & 0.01 \\
Qwen3-VL-2B & Hybrid16 & 0.09 & 0.13 & 0.03 & 0.01 \\
Qwen3-VL-4B & Even16   & 0.15 & 0.21 & 0.06 & 0.02 \\
Qwen3-VL-4B & Hybrid16 & 0.14 & 0.20 & 0.07 & 0.03 \\
Qwen3-VL-8B & Even16   & 0.21 & 0.33 & 0.12 & 0.05 \\
Qwen3-VL-8B & Hybrid16 & 0.22 & 0.36 & 0.13 & 0.04 \\
\midrule
\multicolumn{6}{l}{\textit{Fine-tuned 2B (16 Frames)}} \\
BBx-only    & Even16   & 0.06 & 0.08 & 0.03 & 0.01 \\
BBx-only    & Hybrid16 & 0.06 & 0.09 & 0.05 & 0.02 \\
Even-only   & Even16   & 0.51 & 0.72 & 0.57 & 0.36 \\
Even-only   & Hybrid16 & 0.49 & 0.70 & 0.54 & 0.31 \\
Hybrid-only & Even16   & 0.24 & 0.38 & 0.19 & 0.08 \\
Hybrid-only & Hybrid16 & \textbf{0.73} & 0.81 & 0.77 & 0.68 \\
Evenbbx     & Even16   & 0.47 & 0.70 & 0.52 & 0.30 \\
Evenbbx     & Hybrid16 & 0.43 & 0.61 & 0.46 & 0.27 \\
Hybridbbx   & Even16   & 0.34 & 0.53 & 0.33 & 0.13 \\
Hybridbbx   & Hybrid16 & 0.69 & 0.78 & \textbf{0.74} & \textbf{0.64} \\
\bottomrule
\end{tabular}}
\caption{Complete temporal grounding results across all configurations 
and test distributions.}
\label{tab:full_temporal}
\end{table*}

\begin{table}[htbp]
\centering
\small
\renewcommand{\arraystretch}{1.15}
\begin{tabular}{lccc}
\toprule
\textbf{Configuration} & \textbf{mIoU} & \textbf{P@0.5} & \textbf{P@0.7} \\
\midrule
\multicolumn{4}{l}{\textit{Zero-shot Baselines}} \\
Qwen3-VL-2B & 0.03 & 0.00 & 0.00 \\
Qwen3-VL-4B & 0.03 & 0.00 & 0.00 \\
Qwen3-VL-8B & 0.03 & 0.00 & 0.00 \\
\midrule
\multicolumn{4}{l}{\textit{Fine-tuned 2B}} \\
BBx-only    & \textbf{0.60} & \textbf{0.67} & 0.52 \\
Even-only   & 0.01 & 0.00 & 0.00 \\
Hybrid-only & 0.00 & 0.00 & 0.00 \\
Evenbbx     & 0.55 & 0.61 & 0.44 \\
Hybridbbx   & 0.60 & 0.67 & \textbf{0.47} \\
\bottomrule
\end{tabular}
\caption{Complete spatial grounding results.}
\label{tab:full_spatial}
\end{table}

\section{Model Scaling Analysis}

Table~\ref{tab:model_arch} summarizes architectural differences across 
Qwen3-VL model sizes, and Table~\ref{tab:scaling} presents zero-shot 
performance scaling results.

\begin{table}[htbp]
\centering
\small
\renewcommand{\arraystretch}{1.15}
\begin{tabular}{lccc}
\toprule
\textbf{Model} & \textbf{ViT Layers} & \textbf{LM Layers} & 
\textbf{Params} \\
\midrule
Qwen3-VL-2B & 24 & 28 & $\sim$2.5B \\
Qwen3-VL-4B & 24 & 36 & $\sim$4.5B \\
Qwen3-VL-8B & 27 & 36 & $\sim$8.5B \\
\bottomrule
\end{tabular}
\caption{Qwen3-VL model architecture specifications. The 2B and 4B 
models share identical vision encoders; only the language model depth 
differs.}
\label{tab:model_arch}
\end{table}

\begin{table}[htbp]
\centering
\small
\renewcommand{\arraystretch}{1.15}
\begin{tabular}{lccc}
\toprule
\textbf{Model} & \textbf{Even16} & \textbf{Hybrid16} & 
\textbf{Spatial} \\
\midrule
Qwen3-VL-2B & 0.13 & 0.09 & 0.03 \\
Qwen3-VL-4B & 0.15 & 0.14 & 0.03 \\
Qwen3-VL-8B & 0.21 & 0.22 & 0.03 \\
\midrule
2B$\rightarrow$4B gain & +0.02 & +0.05 & 0.00 \\
4B$\rightarrow$8B gain & +0.07 & +0.08 & +0.01 \\
\bottomrule
\end{tabular}
\caption{Zero-shot performance scaling across model sizes. 
The 2B$\rightarrow$4B transition expands only the language model 
(28$\rightarrow$36 layers); the 4B$\rightarrow$8B transition also 
expands the vision encoder (24$\rightarrow$27 layers), yielding 
larger gains despite fewer added layers.}
\label{tab:scaling}
\end{table}

\section{Spatial Annotation Details}

Table~\ref{tab:categories} shows the top object categories in the 
generated spatial annotations before subsampling. Door and person 
are heavily overrepresented, reflecting the indoor household activity 
focus of Charades-STA. Subsampling to 3,951 annotations preserves 
category diversity while preventing class imbalance during training.

\begin{table}[htbp]
\centering
\small
\renewcommand{\arraystretch}{1.15}
\begin{tabular}{lcc}
\toprule
\textbf{Category} & \textbf{Instances} & \textbf{Percentage} \\
\midrule
door    & 8,586 & 16.2\% \\
person  & 4,718 & 8.9\%  \\
book    & 3,200 & 6.0\%  \\
towel   & 2,100 & 4.0\%  \\
blanket & 1,800 & 3.4\%  \\
\midrule
Total (pre-subsample) & 53,000+ & 100\% \\
Final annotations     & 3,951   & --    \\
\bottomrule
\end{tabular}
\caption{Top object categories in spatial annotations before 
subsampling.}
\label{tab:categories}
\end{table}

\section{Implementation Details}

All models are trained using the LLaMA-Factory framework on a single 
NVIDIA H100 GPU (80GB) with PyTorch 2.9.1 and Hugging Face 
Transformers. Table~\ref{tab:training_hyperparams} provides complete 
hyperparameter specifications.

\begin{table}[t]
\centering
\caption{Training hyperparameters}
\label{tab:training_hyperparams}
\resizebox{\columnwidth}{!}{%
\begin{tabular}{llll}
\toprule
Parameter & Value & Notes & Applies to \\
\midrule
Optimizer & AdamW & -- & All \\
Learning Rate & $5\times10^{-6}$ & -- & All \\
Batch Size (per device) & 4 / 2 & 4 for BBx, 2 for others & BBx / Other \\
Gradient Accumulation Steps & 8 & -- & All \\
Effective Batch Size & 32 / 16 & 32 for BBx, 16 for others & BBx / Other \\
Training Epochs & 3 & -- & All \\
Warmup Ratio & 0.1 & Linear warmup for first 10\% of steps & All \\
LR Scheduler & Cosine & Cosine annealing after warmup & All \\
Mixed Precision & BF16 & -- & All \\
Gradient Checkpointing & Enabled & -- & All \\
Max Sequence Length & 1024 / 4096 & 1024 for BBx, 4096 for others & BBx / Other \\
Image Min Pixels & 65,536 & $256\times256$ & All \\
Image Max Pixels & 262,144 & $512\times512$ & All \\
\bottomrule
\end{tabular}}
\end{table}

\section{Spatial-Temporal Coupling Analysis}

Table~\ref{tab:coupling} illustrates an interaction between spatial 
and temporal grounding on video 50E06, which contains multiple queries 
involving small objects. When spatial localization is accurate (eats 
food, IoU$_s$=0.84), temporal grounding is also strong 
(IoU$_t$=1.00). When spatial accuracy degrades (drinks from cup, 
IoU$_s$=0.45), temporal performance also suffers (IoU$_t$=0.59), 
suggesting the two tasks interact through shared visual representations.

\begin{table}[htbp]
\centering
\small
\renewcommand{\arraystretch}{1.15}
\resizebox{0.7\columnwidth}{!}{%
\begin{tabular}{lcccc}
\toprule
\textbf{Query} & \textbf{GT} & \textbf{Pred} & 
\textbf{IoU\textsubscript{t}} & \textbf{IoU\textsubscript{s}} \\
\midrule
drinks from cup   & 15.9--24.9 & 15.9--21.2 & 0.59 & 0.45 \\
eats food         & 0.0--5.1   & 0.0--5.1   & 1.00 & 0.84 \\
puts cup down     & 21.2--27.1 & 15.9--21.2 & 0.00 & 0.29 \\
\bottomrule
\end{tabular}}
\caption{Spatial-temporal coupling effects (Video 50E06). 
IoU$_t$ = temporal IoU, IoU$_s$ = spatial IoU.}
\label{tab:coupling}
\end{table}

\section{Practical Deployment Scenarios}

We outline three representative deployment scenarios and recommended 
configurations based on our experimental findings.

\paragraph{Video Archive Retrieval.}
For pre-annotated video archives, Hybridbbx with Hybrid16 sampling 
is preferred. Given a query such as ``person drinking coffee,'' the 
system samples 8 frames from the annotated action segment and 8 frames 
uniformly, achieving 68\% temporal mIoU and 60\% spatial mIoU. 
Processing a 30-second video requires approximately 2--3 seconds 
on standard hardware.

\paragraph{Content Moderation.}
Large-scale platforms processing high volumes of user uploads should 
use a two-tiered approach. First, Evenbbx with uniform 16-frame 
sampling provides fast baseline screening at approximately 200ms per 
video, achieving 42--47\% temporal mIoU and 55\% spatial mIoU. 
Videos flagged for potential violations are then reprocessed with 
Hybridbbx, increasing processing time to approximately 3 seconds 
while improving temporal mIoU to 68\%. This cascaded approach 
balances throughput with accuracy for critical cases.

\paragraph{Real-Time Monitoring.}
Systems with strict latency constraints and no temporal annotations 
should use Evenbbx with uniform 16-frame sampling, achieving 
42--47\% temporal mIoU at approximately 150ms per video. New 
cameras can be added without retraining or manual annotation, 
providing operational flexibility.

In all scenarios, fine-tuned 2B models with sparse sampling 
offer the best balance of accuracy, efficiency, and deployability. 
Dense frame processing with large zero-shot models is not recommended 
due to high computational cost without meaningful accuracy gains.

\clearpage
\section{Qualitative Examples}
\label{app:qualitative}

The following figures present representative success and failure cases 
from the Hybridbbx model evaluated on Hybrid16 test data. Each figure 
shows the sampled frames alongside the query, ground truth temporal 
boundaries, and model predictions.

\begin{figure*}[!b]
\centering
\caseimage{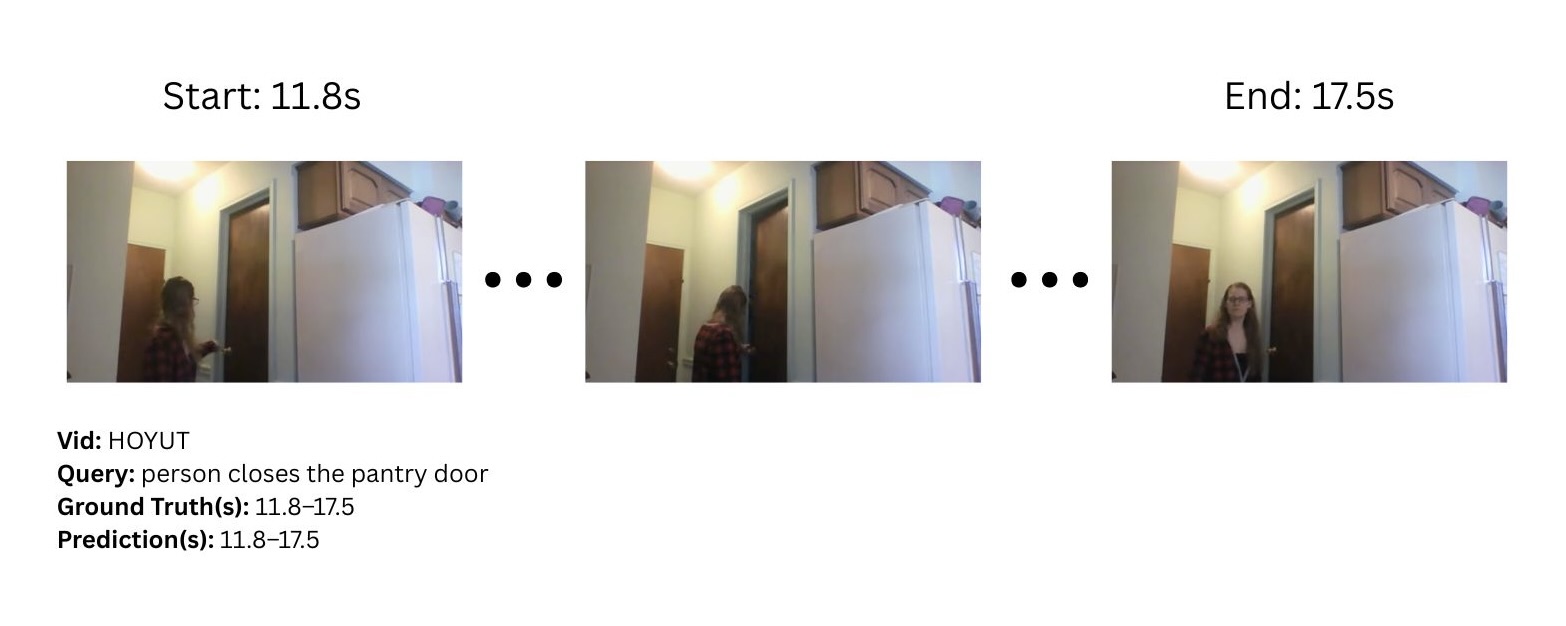}
\vspace{0.35em}
\caseimage{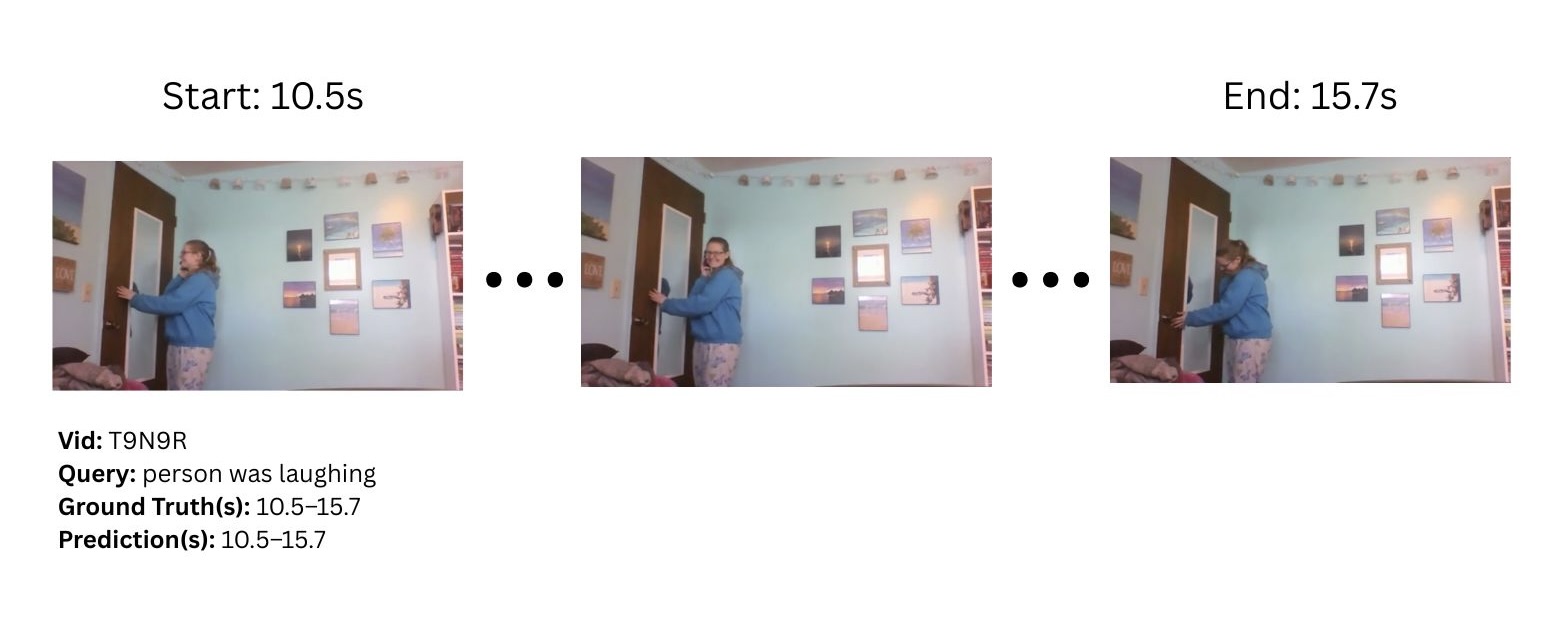}
\vspace{0.35em}
\caseimage{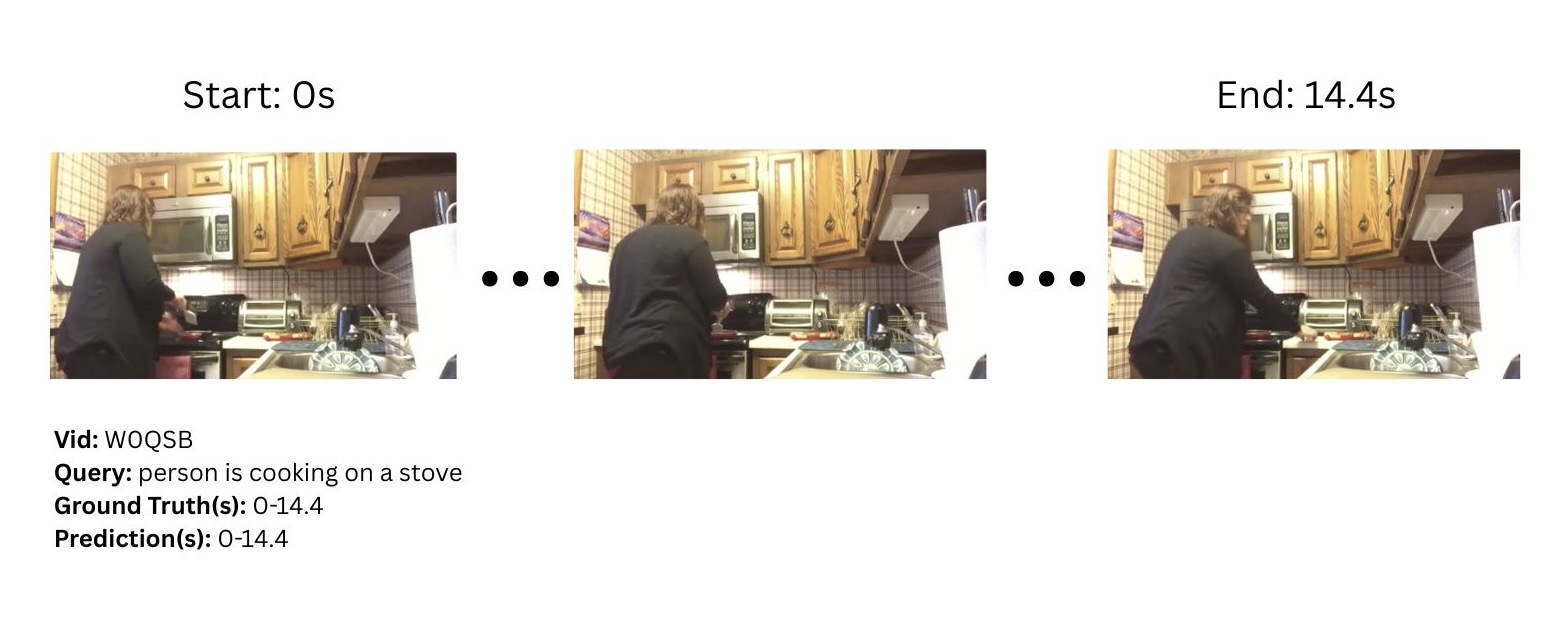}
\caption{Success cases from the Hybridbbx model on 
Hybrid16 test data. Top: \textit{person closes the pantry door} 
(GT: 11.8--17.5s, Pred: 11.8--17.5s, IoU: 1.00). Middle: 
\textit{person was laughing} (GT: 10.5--15.7s, Pred: 10.5--15.7s, 
IoU: 1.00). Bottom: \textit{person is cooking on a stove} 
(GT: 0.0--14.4s, Pred: 0.0--14.4s, IoU: 1.00).}
\end{figure*}

\FloatBarrier

\begin{figure*}[!t]
\centering
\caseimage{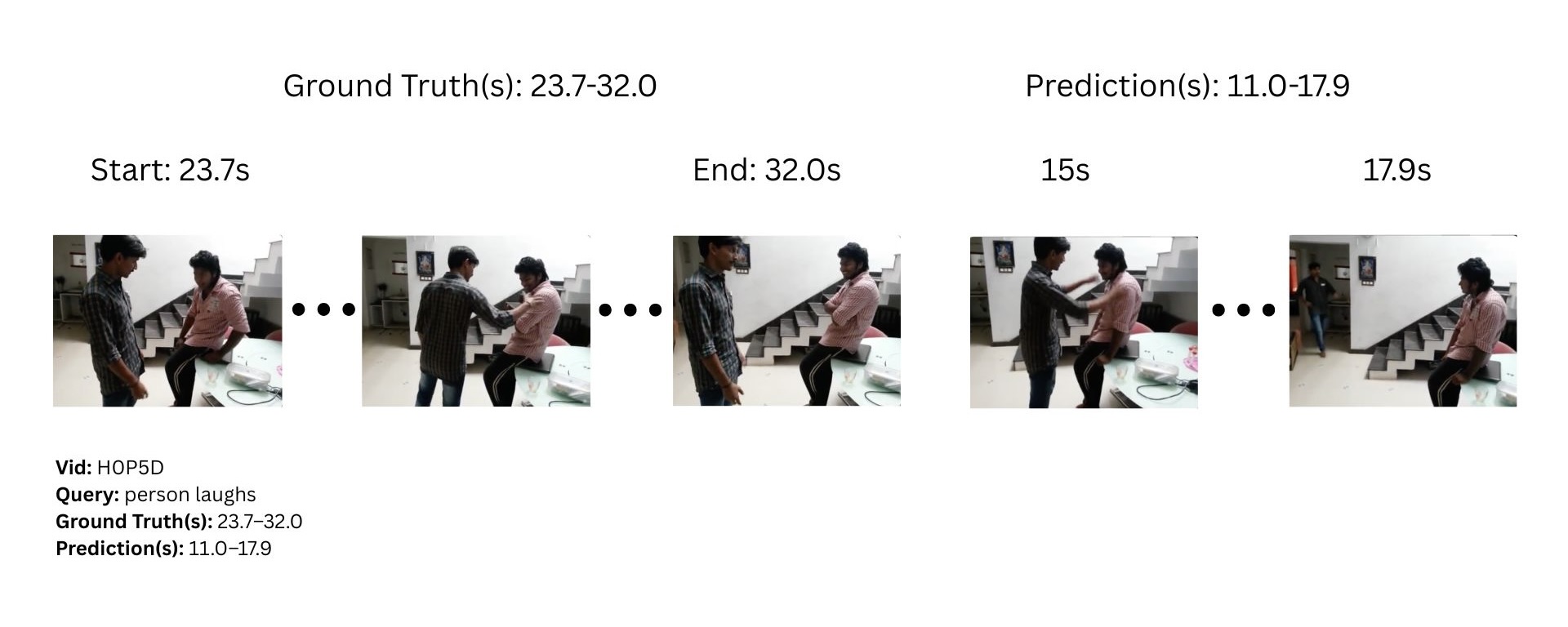}
\vspace{0.35em}
\caseimage{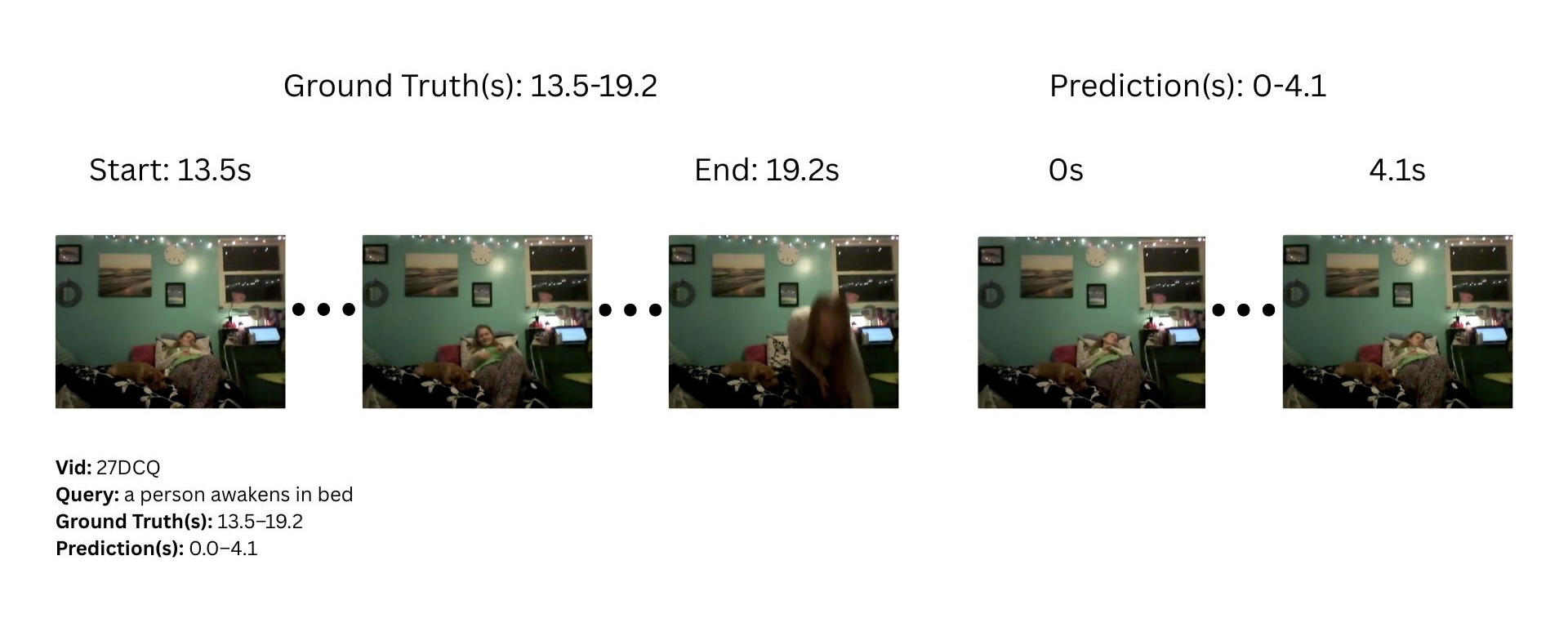}
\vspace{0.35em}
\caseimage{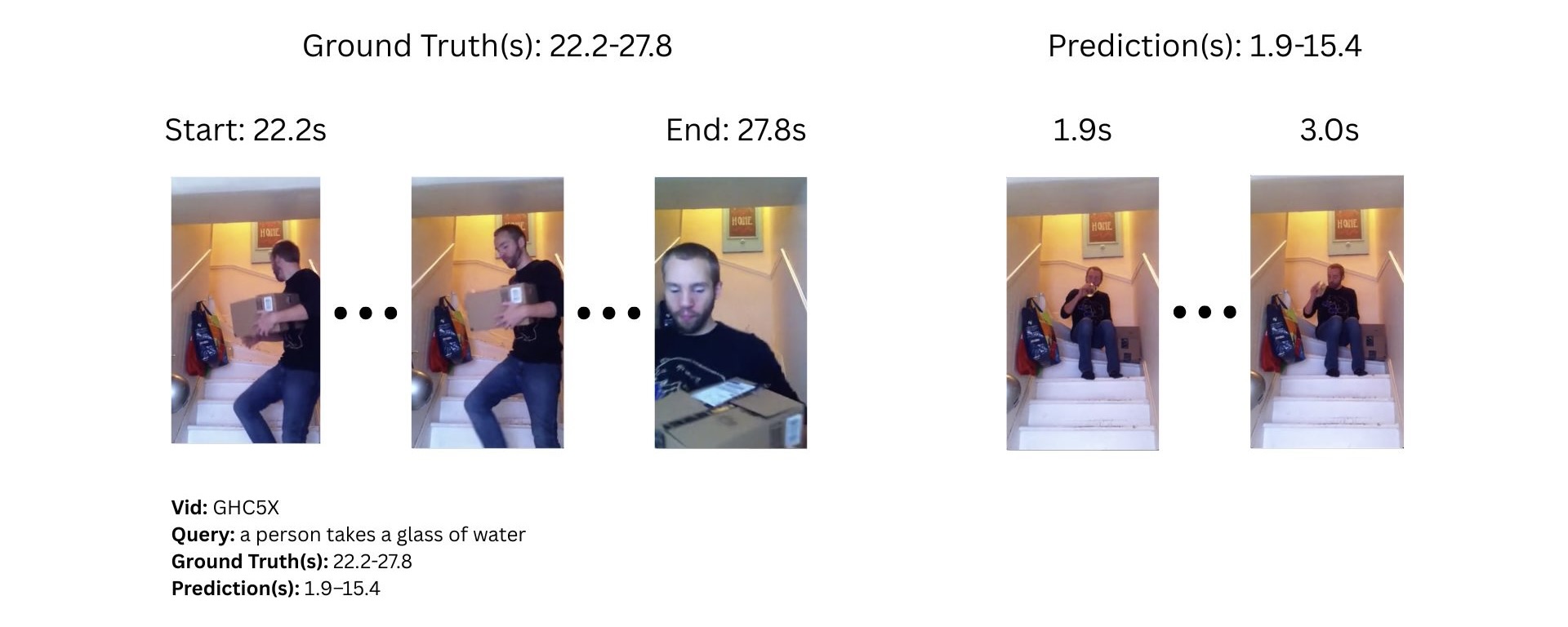}
\caption{Failure cases from the Hybridbbx model on 
Hybrid16 test data. Top: \textit{person laughs} (GT: 23.7--32.0s, 
Pred: 11.0--17.9s, IoU: 0.00), where the model predicts the same 
interval as the co-occurring \textit{person runs into a table} action. 
Middle: \textit{person awakens in bed} (GT: 13.5--19.2s, Pred: 
0.0--4.1s, IoU: 0.00), a gradual state change with minimal discrete 
motion. Bottom: \textit{person takes a glass of water} 
(GT: 22.2--27.8s, Pred: 1.9--15.4s, IoU: 0.00), a small-object 
interaction with a 20-second temporal offset.}
\end{figure*}

\end{document}